\pdfoutput=1

\documentclass[11pt]{article}

\usepackage{emnlp2021}

\usepackage{times}
\usepackage{latexsym}

\usepackage[T1]{fontenc}

\usepackage[utf8]{inputenc}

\usepackage{microtype}

\usepackage{xcolor}
\usepackage{comment}
\usepackage{latexsym}
\usepackage{graphicx}
\usepackage{amsmath}
\usepackage{amssymb}
\usepackage{multirow}
\usepackage{booktabs}
\usepackage{subfigure}
\usepackage{url}
\usepackage{hyperref}
\usepackage{tikz}
\usepackage{tabularx}

\title{\textsc{DiscoDVT}: Generating Long Text with Discourse-Aware Discrete Variational Transformer}

\author{Haozhe Ji$^1$, Minlie Huang$^1$\thanks{\quad Corresponding author}  \\
$^1$Department of Computer Science and Technology,
Institute for Artificial Intelligence, \\
State Key Lab of Intelligent Technology and Systems, \\ 
Beijing National Research Center for Information Science and Technology, \\ 
Tsinghua University, Beijing 100084, China \\
  {\tt\small jhz20@mails.tsinghua.edu.cn,}
  {\tt\small aihuang@tsinghua.edu.cn } \\}
  
\begin{document}
\maketitle
\begin{abstract}
Despite the recent advances in applying pre-trained language models to generate high-quality texts, generating long passages that maintain long-range coherence is yet challenging for these models. In this paper, we propose \textsc{DiscoDVT}, a discourse-aware discrete variational Transformer to tackle the incoherence issue. \textsc{DiscoDVT} learns a discrete variable sequence that summarizes the global structure of the text and then applies it to guide the generation process at each decoding step. To further embed discourse-aware information into the discrete latent representations, we introduce an auxiliary objective to model the discourse relations within the text. We conduct extensive experiments on two open story generation datasets and demonstrate that the latent codes learn meaningful correspondence to the discourse structures that guide the model to generate long texts with better long-range coherence.\footnote{The source code is available at \url{https://github.com/cdjhz/DiscoDVT}.}
\end{abstract}

\section{Introduction}

Generating passages that maintain long-range coherence is a long-standing problem in natural language generation (NLG). 
Despite the recent advances of large pre-trained language generation models~\citep{radford2019language,bart} in various NLG tasks such as summarization and dialogue generation that target generating \textit{locally} coherent texts which are relatively short, it is still challenging for pre-trained models to generate \textit{globally} coherent passages spanning over dozens of sentences.

\begin{figure}[t!]
    \centering
    \includegraphics[width=\columnwidth]{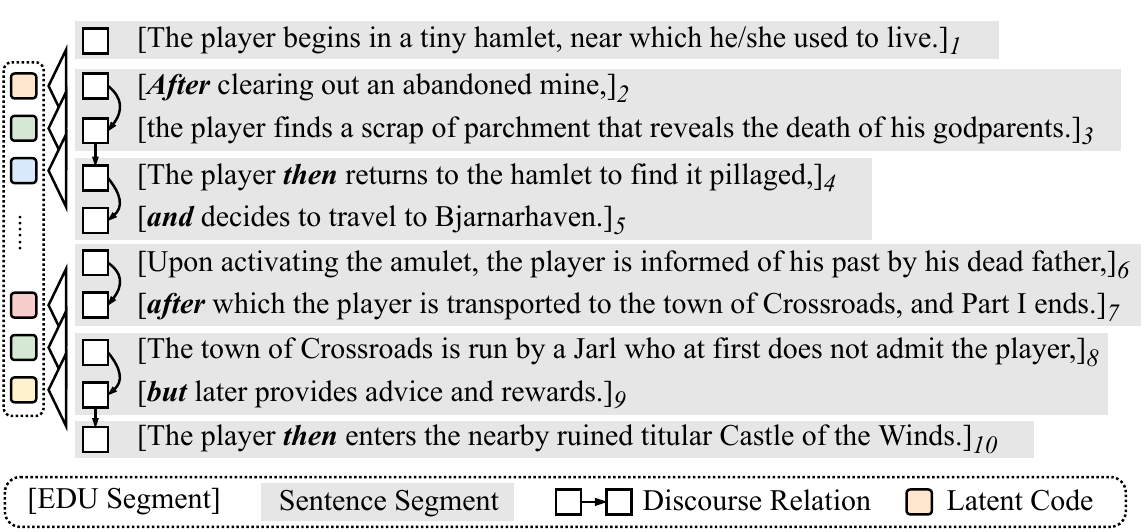}
    \caption{An example from Wikiplots where
    a discrete latent variable sequence abstracts the discourse structure of the text. Inter-sentence and intra-sentence discourse relations are indicated by the \textbf{bolded} discourse markers.}
    \label{fig:example}
    \vspace{-6pt}
\end{figure}

Global coherence in human texts is represented by the topic-maintenance and natural transition between viewpoints~\citep{speech_DanJurafsky}. As illustrated in Figure \ref{fig:example}, discourse relations such as \textit{causal, temporal succession} between contiguous text segments are commonly indicated by the highlighted \textit{discourse markers} which bind collocated text segments into a global structure~\citep{Hobbs1985OnTC}.
Although pre-trained language models are inspected to perform reasonably well in associating topic-related concepts, they can hardly arrange contents with well-structured discourses~\citep{See,ko-li-2020-assessing}.

In this work, we urge the revival of variational autoencoder (VAE) with its global representation ability to tackle the incoherence issue in long text generation in the era of pre-trained language models. To represent texts with high-level structures, we propose to learn a latent variable sequence with each latent code abstracting a local text span.
Instead of the commonly used continuous latent variables, we resort to learn discrete latent codes that naturally correspond to interpretable categories in natural languages~\citep{DBLP:conf/acl/EskenaziLZ18}.
For the latent codes to capture the explicit discourse structure of the texts as shown in Figure \ref{fig:example}, we further design an auxiliary objective on the latent representations to model the discourse relations.


We name the proposed model as \textsc{DiscoDVT}, i.e., \textit{\textbf{Disco}urse-aware \textbf{D}iscrete \textbf{V}ariational \textbf{T}ransformer}. The main idea is to learn a discrete latent variable sequence that summarizes the long text to reconstruct the original text by guiding the decoding process. The learning schema is shown in Figure \ref{fig:model} (a). \textbf{At the encoding phase}, to capture the high-level structure of the text, we first use a bidirectional encoder to obtain contextualized token representations and then apply 1-dimensional convolutional neural networks (1D CNNs) to abstract text segments at the temporal scale (\S{\ref{sec:cnn}}). To condense the continuous representations into categorical features, we map them into a one-hot categorical distribution over a fixed latent vocabulary, and obtain the discrete variable sequence (\S{\ref{sec:bottleneck}}). \textbf{At the decoding phase}, to apply the global discrete latent codes to guide the local text realization, the latent embeddings are first rescaled to the text length with transposed 1D CNNs, and then added to the embedding layer of the decoder for step-wise control (\S{\ref{sec:control}}). For the latent codes to abstract the discourse structure of the text, we use explicit discourse relations from Penn Discourse TreeBank 2.0 (PDTB, \citealp{pdtb2}) and extract adjacent elementary discourse units (EDUs) from texts as shown in Figure \ref{fig:example} and introduce an anxiliary objective to embed the relations into the latent representations.
Once the discrete latent codes are learned, we adopt an autoregressive Transformer to model the prior distribution as a sequence transduction task (\S{\ref{sec:stage2}}).

We summarize our contributions in three folds:

(1) We propose a novel latent variable model that learns discrete latent variable sequence from the long text and applies it to guide the generation process to maintain long-term coherence.

(2) We further acquire the discourse relation information and introduce an auxiliary objective for the discrete latent codes to abstract the discourse structure of the text.

(3) We conduct extensive experiments on two open story generation datasets with automatic and human evaluation. Results demonstrate that our model outperforms baselines in generating coherent long texts with interpretable latent codes.

\section{Related Work}

\subsection{Long Text Generation}
Prior works endeavored to solve the incoherence issue in long text generation can be mainly categorized into model structure modifications, generation mechanism modifications, and prior knowledge injection. 

To model the hierarchical nature of human texts, \citet{JiweiHieRNN} proposed a hierarchical RNN decoder to learn sentence-level representations within the paragraph. \citet{DBLP:conf/acl/ShenCZCWGC19} augmented the hierarchical model with multi-level latent variables, and \citet{DBLP:conf/emnlp/ShaoHWXZ19} further incorporates a planning mechanism to pre-arrange the order and the group of the input keywords. 

Another line of works proposed to decompose long text generation into multiple  stages~\citep{DBLP:conf/acl/LewisDF18,plan-and-write,Fan19,DBLP:journals/corr/abs-2006-15720} where the model first generates a rough sketch, such as key phrases or summaries, and then expands it into the complete long text with fine detail. However, the multi-step generation method is known to have the stage-level exposure bias~\citep{DBLP:journals/corr/abs-2006-15720}, i.e., the discrepancy of middle-stage outputs during training and inference, which can accumulate error through stages and impair the final generation quality. 

The final direction is to inject prior external knowledge into pre-trained language models for commonsense story generation~\citep{GuanTACL,DBLP:conf/emnlp/XuPSPFAC20}. However, these methods may not be generalizable to different data genres such as fictional stories and do not provide long-range guidance during text generation.

\begin{figure*}[t!]
    \centering
    \vspace{-5pt}
    \includegraphics[width=2.0\columnwidth]{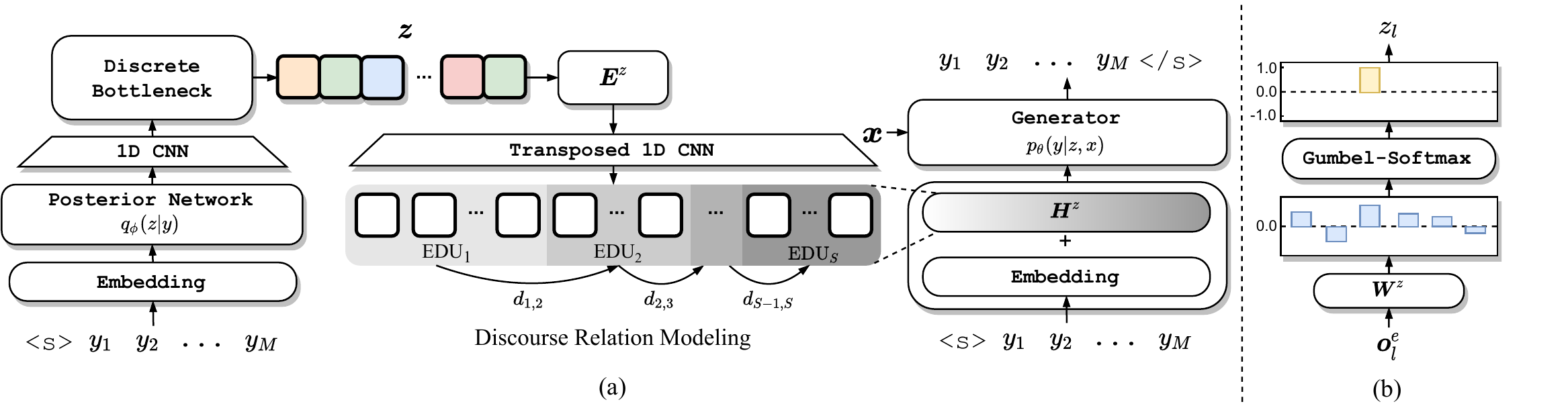}
    \caption{Overview of \textsc{DiscoDVT}. (a) Learning discrete latent codes via encoding and reconstructing the target text with discourse relation modeling where the latent representations are required to predict discourse relations within the text. 
    (b) The discrete variational bottleneck maps the output of the encoder into a categorical distribution over a fixed latent vocabulary.}
    \label{fig:model}
\end{figure*}

\subsection{Discrete Latent Variable Models}

In text generation, continuous Gaussian VAEs have been explored to model response diversity~\citep{DBLP:conf/acl/ZhaoZE17,DBLP:conf/aaai/SerbanSLCPCB17} and high-level structures, such as template and order~\citep{DBLP:conf/emnlp/WisemanSR18,DBLP:conf/acl/ShenCZCWGC19,DBLP:conf/emnlp/ShaoHWXZ19}.
Aside from Gaussian latent variables in the continuous space~\citep{DBLP:journals/corr/KingmaW13}, recent works also explored VAEs in the discrete space~\citep{DBLP:conf/iclr/Rolfe17} with the merit of explainability and revealed the correspondence between the latent codes and categorical features, e.g., dialogue acts~\citep{DBLP:conf/acl/EskenaziLZ18} and POS tags~\citep{DBLP:journals/corr/abs-2103-11405}.
Recently, \citet{DBLP:conf/nips/OordVK17} proposed a vector-quantized variational autoencoder (VQ-VAE) which circumvents the posterior collapse problem by learning quantized one-hot posterior that can be adapted with powerful autoregressive decoders. 
In image and speech generation, \citet{vq-vae-2,DBLP:conf/nips/DielemanOS18} extended VQ-VAE with hierarchical latent variables to capture different input data resolutions and generate high-fidelity visual and audio data with high-level structures.
To our knowledge, in the domain of text generation, our work is the first attempt that explores discrete latent variable models scaling up to the size of large pre-trained language models to solve the incoherence issue in long text generation.

\section{Methodology}

\subsection{Task Definition and Model Overview}

We formulate the long text generation task as a conditional generation problem, i.e., generating a multi-sentence text $\boldsymbol{y}=(y_1,\cdots,y_M)$ given an input prompt $\boldsymbol{x}=(x_1,\cdots,x_N)$. Current pre-trained generation models, e.g., BART, adopt Transformer-based encoder-decoder structure that bidirectionally encodes $\boldsymbol{x}$ and maximizes the log-likelihood $\mathcal{L}_{\text{LM}}$ of predicting $\boldsymbol{y}$ at the decoder side.

However, existing models can hardly maintain long-range coherence when generating long texts that span hundreds of words. We propose to learn a discrete sequence of latent variables $\boldsymbol{z}=(z_1,\cdots, z_L)$ to abstract the high-level structure of the text at the temporal scale ($L$ is much shorter than $M$) and categories (each $z_l$ takes value from the latent vocabulary with size $K$, which is much smaller than the text vocabulary).


Our model maximizes the evidence lower bound (ELBO, ~\citealp{DBLP:journals/corr/KingmaW13}) of the log-likelihood of
the generative model: $p(\boldsymbol{y},\boldsymbol{z}|\boldsymbol{x})=p_\theta(\boldsymbol{y}|\boldsymbol{z},\boldsymbol{x})p_\psi(\boldsymbol{z}|\boldsymbol{x})$ where the \textit{generator} and the \textit{prior network} are parametrized by $\theta$ and $\psi$, respectively. 
Since we want $\boldsymbol{z}$ to capture the internal structure of the text instead of specific topic information, we posit it to be independent of the input prompt $\boldsymbol{x}$ and formulate the \textit{posterior network} as $q_\phi(\boldsymbol{z}|\boldsymbol{y})$. The same formulation is also adopted by \citet{DBLP:conf/acl/EskenaziLZ18} to learn interpretable latent variables.
We give the ELBO in the following:
\begin{align}\label{equ:elbo}
    \mathcal{L}_{\text{ELBO}} = \mathbb{E}_{\boldsymbol{z}\sim q_\phi} & \log p_\theta(\boldsymbol{y}|\boldsymbol{z},\boldsymbol{x}) - \nonumber\\
    & D_{\text{KL}}(q_\phi(\boldsymbol{z}|\boldsymbol{y})\|p_\psi(\boldsymbol{z}|\boldsymbol{x})).
\end{align}

Due to the discrete nature of $\boldsymbol{z}$, $q_\phi(\boldsymbol{z}|\boldsymbol{y})$ defines a sequence of one-hot distribution over the discrete vocabulary $K$ at each position. Thus, the second term of the ELBO can be interpreted as a sequence transduction objective that autoregressively fits the prior model to the target sequence $\boldsymbol{z}$ given by the posterior: $p_\psi(\boldsymbol{z}|\boldsymbol{x})=\prod_l p_\psi(z_l|\boldsymbol{z}_{<l},\boldsymbol{x})$. 

We follow \citet{DBLP:conf/nips/OordVK17} and separate the learning process into two stages. In the first training stage, we train the \textit{posterior network} and the \textit{generator} to optimize the first term of the ELBO to learn discrete latent codes of the text (\S{\ref{sec:stage1}}). We further propose a discourse-aware objective for the latent representations to model high-level discourse relations of the text (\S{\ref{sec:disc}}). In the second training stage, we adopt another Transformer model as the \textit{prior network} that predicts the discrete latent codes given the input prompt (\S{\ref{sec:stage2}}). During the inference stage, we first sample a sequence of latent variables from the prior network given the input prompt, and then inject it into the generator to guide the local text realization by randomly sampling text tokens.

\subsection{Learning Discrete Latent Codes}\label{sec:stage1}

In this section, we introduce the procedure of learning discrete latent codes from the long text. Given the text $\boldsymbol{y}$, the idea is to encode it into a latent variable sequence $\boldsymbol{z}$ that preserves high-level structure to guide text reconstruction with the input $\boldsymbol{x}$. $\boldsymbol{y}$ is first encoded into contextualized representations with a bidirectional Transformer encoder, and then abstracted into $\boldsymbol{z}$ with 1D CNNs and the discrete variational bottleneck. To guide text generation, we first embed $\boldsymbol{z}$ into the embedding matrix, then rescale it to the original length of $\boldsymbol{y}$ with transposed 1D CNNs, and finally inject it into the decoder's embedding layer for step-wise control.

\subsubsection{Temporal Abstraction with CNNs}\label{sec:cnn}

To abstract high-level features that correspond to the global structure of the text, we adopt $c$-layer 1D CNNs that decrease the text length $2^c$ times where each layer halves the input size. The similar architecture was also explored in non-autoregressive machine translation~\citep{DBLP:conf/icml/KaiserBRVPUS18} but for the purpose of parallel decoding.

Formally, given the input text representations $\boldsymbol{H}^e=[\boldsymbol{h}_1^e,\cdots,\boldsymbol{h}_M^e]$, the output of CNNs is denoted as $\boldsymbol{O}^e=[\boldsymbol{o}_1^e,\cdots,\boldsymbol{o}_L^e]$. Intuitively, stacked CNNs extract contiguous $n$-gram features from the text sequence with each code abstracting a contiguous text span with flexible boundaries.

At the decoding phase, to smoothen the high-level representations at the temporal level for continuous local text generation, we adopt transposed CNNs with the symmetric structure to rescale the code embedding matrix $\boldsymbol{O}^z=[\boldsymbol{o}_1^z,\cdots,\boldsymbol{o}_L^z]$ into low-level features $\boldsymbol{H}^z=[\boldsymbol{h}_1^z,\cdots,\boldsymbol{h}_M^z]$.


\subsubsection{Discrete Variational Bottleneck}\label{sec:bottleneck}

To enforce $\boldsymbol{z}$ to preserve salient information for text reconstruction with interpretable categories, we introduce a discrete variational bottleneck that discretizes the 
CNN outputs 
$\boldsymbol{O}^e$ into categorical features. Intuitively, the bottleneck controls the information capacity of $\boldsymbol{z}$ by mapping continuous representations to a discrete space.

We give a formal description of the discretization. Figure\ref{fig:model} (b) presents an example at the $l$-th position\footnote{We ommit the subscript $l$ in the following derivation.}. ${\boldsymbol{o}}^e$ is first mapped into logits $\boldsymbol{t}=\boldsymbol{W}^z{\boldsymbol{o}}^e\in \mathbb{R}^K$ through a linear transformation. The discrete code $z$ at this position is defined as
\begin{equation}
    z = \underset{k\in K}{\text{argmax}} \ t_k.
\end{equation}
During training, to backpropagate gradients, we apply the Gumbel-Softmax trick~\citep{DBLP:conf/iclr/JangGP17,DBLP:conf/iclr/MaddisonMT17} to provide a differentiable relaxation of the argmax operation.
\begin{equation}
    w_k = \frac{\exp((t_k + g_k)/\tau)}{\sum_{k=1}^K \exp((t_k + g_k)/\tau)},
\end{equation}
where $g_1, \cdots, g_K$ are i.i.d samples from the Gumbel distribution, and $\tau$ is the temperature that controls the tightness of the relaxation. As $\tau$ anneals from $\tau_{max}$ to nearly 0 during training, the soft categorical distribution $w_k$ becomes a reasonable estimation of the one-hot distribution. 

This categorical distribution is then multiplied to the learnable code embeddings $\boldsymbol{E}^z$ to obtain the code embedding matrix ${\boldsymbol{o}}^z = \boldsymbol{E}^z\boldsymbol{w}$. 

\subsubsection{Generation with Step-Wise Control}\label{sec:control}

For the high-level latent codes to explicitly guide the local text realization, the code embedding matrix $\boldsymbol{O}^z$ is first rescaled into $\boldsymbol{H}^z$. It is then added to the decoder's input embedding layer with token embeddings $\{\boldsymbol{e}_m\}_{m=1}^M$ and positional encodings $\{\boldsymbol{p}_m\}_{m=1}^M$ at each decoding position. The new input embeddings are $\{\boldsymbol{h}^z_m + \boldsymbol{e}_m + \boldsymbol{p}_m\}_{m=1}^M$. Because of the residual structure in the Transformer, the information of $\boldsymbol{h}_m^z$ can be effectively transmit to the higher layers with positional awareness.

Intuitively, each latent code controls the detailed generation of a local text span while different codes summarize diverse high-level patterns in the text.

The reconstruction goal is thus to maximize the following expectation of log-likelihood:
\begin{equation}
    \mathcal{L}_{\text{recon}} = \mathbb{E}_{\boldsymbol{z}\sim q_\phi(\boldsymbol{z}|\boldsymbol{y})}\log p_\theta(\boldsymbol{y}|\boldsymbol{z}, \boldsymbol{x}).
\end{equation}

\subsection{Discourse Relation Modeling}\label{sec:disc}

In order to abstract the discourse structure of the text into the latent representations, we design an auxiliary discourse-aware objective to embed the discourse relation information into the discrete latent codes. We focus on explicit discourse relations rather than implicit discourse signals, e.g., sentence order~\citep{DBLP:conf/naacl/BosselutCHGHC18}, for they cannot express the canonical ways adjacent sentences linked together. 
We select a set of unambiguous discourse markers $\mathcal{D}$ 
from PDTB~\citep{pdtb2} which indicate high-level discourse coherence. As suggested by \citet{pdtb2}, about 90\% of explicit discourse relations appear either in the same sentence or between adjacent sentences.
Thus, for intra-sentence relations, we parse the sentence and extract adjacent EDUs with connected discourse markers based on appropriate dependency patterns following \citet{DisSent}.
The processing details and annotation examples are provided in the \S{\ref{sec:annotation}}.


The discourse annotation results of a single passage are formalized as follows:
\begin{equation}
    \mathcal{A} = \{(s_i,e_i), d_{i,i+1}\}_{i=1}^{S},
\end{equation}
where $S$ is the total number of EDUs in $\boldsymbol{y}$,
$s_i$/$e_i$ are the start/end position of the $i$-th EDU, and $d_{i,i+1}$ is the discourse label between the $i$-th and $i+1$-th EDUs. 

Next, we derive the discourse relation modeling objective formally.
We first obtain the averaged latent representation of the $i$-th EDU $\bar{\boldsymbol{h}}_i$ by mean-pooling the corresponding latent embeddings $[\boldsymbol{h}^z_{s_i},\cdots, \boldsymbol{h}^z_{e_i}]$. Then we use bi-affine transformation to model the relation between two adjacent representations and maximize the log probability as follows:
\begin{align}
    &p(d_{i,i+1}|\boldsymbol{z}) = \text{softmax}(\bar{\boldsymbol{h}}_i^\top \boldsymbol{W}_{d} \bar{\boldsymbol{h}}_{i+1} + \boldsymbol{b}_d),\\
    &\mathcal{L}_{\text{disc}} = \mathbb{E}_{\boldsymbol{z}\sim q_\phi(\boldsymbol{z}|\boldsymbol{y})} \sum_{i=1}^{|\mathcal{A}|-1} \log p(d_{i,i+1}|\boldsymbol{z}).
\end{align}

\subsection{Autoregressive Prior Modeling}\label{sec:stage2}

In the second stage, we propose to use a Transformer encoder-decoder to learn the prior distribution of the discrete latent codes given the input prompt by minimizing the KL divergence term in Eq.(\ref{equ:elbo}) with respect to $\psi$. To facilitate training, we utilize the parameters of a pre-trained text encoder to initialize the encoder of the prior model and train the decoder from scratch. The optimization objective is equivalent to maximize the following log-likelihood.
\begin{equation}
    \mathcal{L}_{\text{prior}} = \mathbb{E}_{\boldsymbol{z}\sim q_\phi(\boldsymbol{z}|\boldsymbol{y})} \sum_{l=1}^L \log p({z}_l|\boldsymbol{z}_{<l}, \boldsymbol{x}).
\end{equation}
In practice, we approximate the expectation by taking argmax from $q_\phi$. Compared to sampling, this approach reduces the learning variance.


\subsection{Additional Learning Techniques}

In preliminary experiments, we found two additional techniques that essentially guide the model to learn meaningful latent abstraction of the text, described below.

\noindent\textbf{Entropy Regularization.} 
We discover that the pre-trained decoder tends to utilize very few discrete codes from the whole code vocabulary, which undermines the expressiveness of the discrete bottleneck. To ensure the model uses the full capacity of the discrete bottleneck, we add an entropy-based regularization that encourages the diverse selection of discrete latent codes across time steps. Specifically, we calculate the average categorical distribution $\bar{\boldsymbol{p}}=\frac{1}{L}\sum_{l=1}^L\text{softmax}(\boldsymbol{t}_l)$ across time steps where $\boldsymbol{t}_l$ is the code logits at the position $l$. Then, we maximize the entropy of the average distribution:
\begin{equation}
    \mathcal{L}_{\text{entr}} = - \sum_{k=1}^K \bar{p}_k\log \bar{p}_k.
\end{equation}

The overall objective to be maximized in the first stage is the weighted sum of the aforementioned objectives: $\mathcal{L}_{\text{recon}} + \lambda_1 \mathcal{L}_{\text{entr}} + \lambda_2 \mathcal{L}_{\text{disc}}$.

\noindent\textbf{Warm-Start Training.} At the beginning of training, if the discrete bottleneck does not produce meaningful latent embeddings, the pre-trained generator will regard them as injected noise, which degrades the generation performance on the downstream tasks. To mitigate this issue, we fix the Gumbel temperature to $\tau_{max}$ and warm-start the model on contiguous texts collected from BookCorpus~\citep{bookcorpus} by maximizing the following objective: $\mathcal{L}_{\text{recon}} + \lambda_1 \mathcal{L}_{\text{entr}}$.

\section{Experiments}

\subsection{Datasets}

We evaluate our model on two open story generation datasets, WritingPrompts and Wikiplots. WritingPrompts~\citep{DBLP:conf/acl/LewisDF18} is a story generation dataset collected from Reddit where users compose fictional stories inspired by short story prompts. WikiPlots\footnote{\url{www.github.com/markriedl/WikiPlots}} corpus contains story plots of various genres, e.g., movies, novels, which are extracted from Wikipedia with story titles. The data statistics are shown in Table \ref{tab:stats}. More details of data processing are provided in \S{\ref{sec:data_proc}}.

\begin{table}[t!]
    \centering
    \scriptsize	
    \begin{tabular}{l|c|c|ccc}
    \toprule[1pt]
         Dataset & Input len. & Output len. & Train & Val & Test\\
    \midrule[0.5pt]
    WritingPrompts & 28.4 & 674.6 & 273K & 15K & 15K\\
    Wikiplots & 3.4 & 354.8 & 101K & 5K & 5K\\
    \bottomrule[1pt]
    \end{tabular}
    \caption{Data statistics of WritingPrompts and Wikiplots include average input/output length and the number of examples in each data split.}
    \label{tab:stats}
\end{table}

\subsection{Implementation Settings}

We utilize the state-of-the-art pre-trained text generation model BART to initialize the components of our model, including the posterior encoder, prior encoder, and the generator. Due to limited computational resources, we use the pre-trained checkpoint of $\text{BART}_{\text{base}}$ for our model and other pre-trained baselines we implemented. We use $3$-layer 1D CNNs with kernel size 4, stride 2, and 0s padding on both sides that downsamples the text sequence into $8$ times shorter discrete latent codes. We set the latent vocabulary size $K=256$ as a tradeoff of latent capacity and computational overhead. In preliminary studies, we found that further increasing the latent vocabulary size requires a longer time to converge while receives little pay back in text diversity and quality. We collect 322K contiguous texts from BookCorpus for warm-start training. We anneal the Gumbel temperature from $\tau_{max}=0.9$ to $\tau_{min}=0.1$ in the first 20K steps during fine-tuning. We set $\lambda_1=0.1, \lambda_2=0.1$. We adopt AdamW~\citep{DBLP:conf/iclr/LoshchilovH19} as the optimizer. More training details are provided in the \S{\ref{sec:train_detail}}.

During inference, we randomly sample 1,000 prompts from each test set for automatic evaluation. 
We use nucleus sampling~\citep{DBLP:conf/iclr/HoltzmanBDFC20} with $p=0.9$, a temperature of $1.0$, and a minimum sequence length of 100 subwords. The same inference settings are applied to all the baselines for fair comparisons.

\subsection{Baselines}

We compare our model to the following baselines:

\noindent\textbf{Seq2Seq} is a Transformer-based sequence-to-sequence model which adopts the same architecture as BART without the pre-trained parameters.

\noindent\textbf{BART}~\citep{bart} is implemented by directly fine-tuning the pre-trained BART model on the downstream datasets.

\noindent\textbf{BART-LM} is implemented by first post-training BART on BookCorpus with the language modeling objective for the same number of steps as ours and then fine-tuning on the downstream datasets. This baseline is proposed to investigate the side effect of the language modeling objective on the decoder in the warm-start stage.

\noindent\textbf{BART-CVAE} is inspired by recent literature that incorporates continuous latent variables to large pre-trained models~\citep{optimus}, which serves as a counterpart to our discrete variable model. We implement a CVAE with modules initialized by the pre-trained parameters of BART. The sampled latent variable is added to the embedding layer of the generator's decoder as our model (same at every position). We adopt the KL thresholding strategy~\citep{NIPS2016_ddeebdee} that maximizes the KL term with a constant $\beta=0.1$ to mitigate the posterior collapse issue.

\noindent\textbf{Aristotelian Rescoring (AR)} is a recent work that incorporates content-planning in BART on WritingPrompts~\citep{DBLP:conf/emnlp/Goldfarb-Tarrant20}. It first generates an SRL-based plot given the prompt and then revises the plot with several rescorers inspired by Aristotle's writing principle and finally generates the long text based on the plot and the prompt. We keep the original model configurations in the paper that adopt $\text{BART}_{\text{large}}$ for text generation and $\text{RoBERTa}_{\text{large}}$ for plot rescoring.

\subsection{Automatic Evaluation}

\textbf{Evaluation Metrics.}
We adopt the following automatic metrics to evaluate the generated stories in terms of (1) relevance, (2) diversity and (3) repetition. (1-a) \textbf{BLEU (B-n)} measures the precision of $n$-grams of the generated texts which are present in the references~\citep{Papineni2001BleuAM}. (1-b) \textbf{MS-Jaccard (MSJ-n)} measures the similarity between the model distribution and the real data distribution by the Jaccard Index between two multi-sets of $n$-grams~\citep{DBLP:journals/corr/abs-1904-03971}. (2-a) \textbf{reverse-BLEU (rB-n)} measures the recall of generated $n$-grams which reflects the diversity of the generation results~\citep{DBLP:conf/ijcai/ShiCQH18}. (2-b) \textbf{Distinct (D-n)} measures the fraction of unique $n$-grams among all the generated $n$-grams~\citep{DBLP:conf/naacl/LiGBGD16}. (3) \textbf{Token Repetition (rep-$l$)} calculates the fraction of the identical token that occurs in the previous $l$ tokens~\citep{DBLP:conf/iclr/WelleckKRDCW20}.

\begin{table*}[t!]
    \centering
    \scriptsize	
    \begin{tabular}{l | ll | ll | ll | ll | ll}
    \toprule[1pt]
    Models & B-1$\uparrow$ & B-2$\uparrow$ & MSJ-2$\uparrow$ & MSJ-3$\uparrow$ & rB-1$\uparrow$ & rB-2$\uparrow$ & D-4$\uparrow$ & D-5$\uparrow$ & rep-8$\downarrow$ & rep-16$\downarrow$ \\
    \midrule[0.5pt]
    \multicolumn{11}{c}{\scriptsize Dataset: \textit{Wikiplots}}\\
    \midrule[0.5pt]
    Seq2Seq & 13.72 & 05.66 & 25.75 & 17.48 & 18.14 & 07.75 & 69.76 & 91.59 & 10.55 & 22.88 \\
    BART & 16.67 & 06.78 & 35.07 & 23.49 & 19.63 & 08.20 & 86.04 & 96.23 & 08.69 & 19.88 \\
    BART-LM & 17.63 & 07.24 & 36.86 & 24.55 & 20.36 & 08.56 & 84.91 & 95.84 & 08.69 & 19.86 \\
    BART-CVAE & 18.16 & 06.74 & 31.35 & 19.37 & 20.31 & 07.74 & \textbf{91.45} & \textbf{98.39} & 11.00 & 22.48 \\
    \midrule[0.5pt]
    \textsc{DiscoDVT} & \textbf{20.57}** & \textbf{08.39}** & \textbf{42.48}** & \textbf{27.38}** & \textbf{22.34}** & \textbf{09.30}** & 90.85 & 98.08 & 
    \textbf{07.50}** & \textbf{17.34}** \\
    \midrule[0.5pt]
    \multicolumn{11}{c}{\scriptsize Dataset: \textit{WritingPrompts}}\\
    \midrule[0.5pt]
    Seq2Seq & 20.93 & 09.03 & 42.02 & 30.55 & 24.52 & 10.85 & 72.04 & 90.71 & 10.86 & 23.46 \\
    BART & 21.64 & 09.41 & 45.07 & 32.31 & 24.95 & {11.04} & 77.99 & 92.32 & 09.54 & 21.70 \\
    BART-LM & 21.76 & 09.43 & 45.46 & 32.50 & 24.91 & {11.01} & 77.65 & 92.20 & 09.48 & 21.53 \\
    BART-CVAE & 20.96 & 07.94 & 33.32 & 21.38 & 22.80 & 08.82 & \textbf{88.14} & \textbf{97.31} & 14.01 & 25.52 \\
    AR & 20.95 & 08.29 & 43.70 & 28.64 & 22.48 & 09.06 & 82.91 & 92.40 & 12.74 & 22.19 \\
    \midrule[0.5pt]
    \textsc{DiscoDVT} & \textbf{24.10}** & \textbf{10.16}** & \textbf{50.00}** & \textbf{34.76}** & \textbf{26.26}** & \textbf{11.29}* & 84.66 & 96.00 & \textbf{09.03}** & \textbf{19.74}** \\
    
    \bottomrule[1pt]
    \end{tabular}
    \caption{Automatic evaluation results on WritingPrompts and Wikiplots. $\uparrow$/$\downarrow$ indicates the higher/lower score, the better. 
    Scores marked with * and ** indicate a significance of $p<0.05$ and $p<0.01$ in the t-test respectively.}
    \label{tab:main_results}
\end{table*}

\noindent\textbf{Results Analysis.} We show the automatic evaluation results on WritingPrompts and Wikiplots in Table \ref{tab:main_results}. 
By comparing \textsc{DiscoDVT} to other baselines, we have the following observations.

On both datasets, \textsc{DiscoDVT} outperforms all the baselines in generating texts with higher $n$-gram overlaps and $n$-gram distribution similarity to the reference texts indicated by a higher BLEU and MSJ score, respectively. 


In terms of diversity, \textsc{DiscoDVT} outperforms BART and BART-LM by large margins in terms of Distinct while slightly underperforms BART-CVAE.
However, when evaluating diversity jointly with quality, \textsc{DiscoDVT} surpasses BART-CVAE with higher reverse-BLEU score. We further examine the generated examples of BART-CVAE and found that its diversity mainly comes from generating more spurious combinations of words that do not appear in the references. This is also evident in the drastic performance drop of the MSJ score.

To quantitatively evaluate the repetition problem of the generated texts, we calculate the token repetition ratio in two different ranges. The results show that \textsc{DiscoDVT} consistently outperforms all the baselines in generating texts with less repetition in both local sentences and contents in a longer range.

Compared to other baselines, AR achieves higher diversity but underperforms in other reference-based metrics. We conjecture that the multi-step scorer suffers from the stage-level exposure bias~\citep{DBLP:journals/corr/abs-2006-15720} that may impair the generation performance.

\subsection{Ablation Study}

We first show the ablation study of different training objectives in Table \ref{tab:ablation}. We first observe a certain performance drop when removing
$\mathcal{L}_{\text{disc}}$ or $\mathcal{L}_{\text{entr}}$
during fine-tuning. Since existing automatic metrics cannot evaluate the discourse structure of texts, we further present a discourse-level evaluation to emphasize the effectiveness of $\mathcal{L}_{\text{disc}}$ in Section \ref{sec:disc_eval}. Then we highlight the significance of warm-start training to learn meaningful latent codes, as the model only uses few latent codes and degenerates to the vanilla BART model when removing it.
We also alter the number of CNN layers and analyze the distribution of code utilization and the generation performance in the \S{\ref{sec:cnn_ablation}}. 

\begin{table}[t!]
    \centering
    \scriptsize
    \begin{tabular}{l | cccc c}
    \toprule[1pt]
    Models & B-1$\uparrow$ & MSJ-2$\uparrow$ & D-4$\uparrow$ & rep-8$\downarrow$\\
    \midrule[0.5pt]
    \textsc{DiscoDVT} & \textbf{20.57} & \textbf{42.48} & \textbf{90.85} & \textbf{07.50} \\
    \ w/o $\mathcal{L}_{\text{disc}}$ & 19.47 & 40.80 & 89.97 & 07.70 \\
    \ w/o $\mathcal{L}_{\text{entr}}$ & 19.30 & 40.15 & 89.68 & 07.84 \\
    \ w/o Warm-start & 18.22 & 35.48 & 90.05 & 09.71 \\
    \bottomrule[1pt]
    \end{tabular}
    \caption{Ablation study of different training objectives on Wikiplots.}
    \label{tab:ablation}
\end{table}

\subsection{Human Evaluation}

For human evaluation, we perform pair-wise comparisons with three strong baselines based on BART. We choose the Wikiplots dataset on which annotators could reach an acceptable agreement given the relatively shorter passage length. We randomly sample 100 prompts from the test set of Wikiplots and obtain the generated texts from the three baselines and ours, resulting in 400 texts in total. We hired three annotators from Amazon Mechanical Turk to give a preference (win, lose, or tie) in terms of coherence and diversity independently. \textit{Coherence} measures whether the story stays on topic and is well-structured with correct logical, temporal, and causal relations. \textit{Informativeness} measures whether the story contains informative details and is engaging on the whole. The final decisions are made by majority voting among the annotators. As shown in Table \ref{tab:human}, \textsc{DiscoDVT} significantly outperforms baselines in both coherence and informativeness, demonstrating that \textsc{DiscoDVT} effectively captures the high-level discourse structure to guide local text realization with details. The results show \textit{moderate} inter-annotator agreement ($0.4\le \kappa < 0.6$).

\begin{table}[t!]
    \centering
    \scriptsize
    \begin{tabular}{l| lll | c}
    \toprule[1pt]
    \multirow{2}{*}{Models} & \multicolumn{4}{c}{Coherence}\\
     & Win & Lose & Tie & $\kappa$ \\
    \midrule[0.5pt]
    \textsc{DiscoDVT} vs. BART & 0.53** & 0.35 & 0.12 & 0.40 \\
    \textsc{DiscoDVT} vs. BART-LM & 0.54** & 0.40 & 0.06 & 0.42 \\
    \textsc{DiscoDVT} vs. BART-CVAE & 0.53** & 0.34 & 0.14 & 0.43 \\
    \midrule[0.5pt]
    \midrule[0.5pt]
    \multirow{2}{*}{Models} & \multicolumn{4}{c}{Informativeness}\\
     & Win & Lose & Tie & $\kappa$ \\
    \midrule[0.5pt]
    \textsc{DiscoDVT} vs. BART & 0.52** & 0.36 & 0.12 & 0.42 \\
    \textsc{DiscoDVT} vs. BART-LM & 0.49* & 0.39 & 0.11 & 0.47 \\
    \textsc{DiscoDVT} vs. BART-CVAE & 0.53** & 0.38 & 0.09 & 0.42 \\
    \bottomrule[1pt]
    \end{tabular}
    \caption{Human evaluation results on Wikiplots. Scores indicate the percentage of Win, Lose, or Tie when comparing \textsc{DiscoDVT} with a baseline. $\kappa$ denotes Fleiss' kappa~\citep{Fleiss1971MeasuringNS}, which measures the inter-annotator agreement. Scores marked with * and ** denote significant differences with $p<0.05$ and $p<0.01$ (sign test) respectively.}
    \label{tab:human}
    \vspace{-10pt}
\end{table}

\begin{figure*}[t!]
    \centering
    \includegraphics[width=2.0\columnwidth]{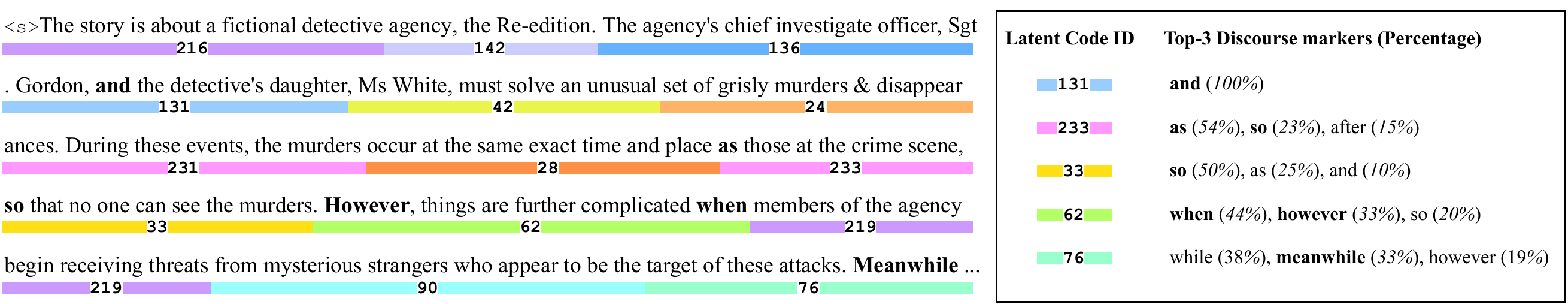}
    \caption{A generated example of \textsc{DiscoDVT} with correspondent latent codes intuitively assigned to text segments of 8 bpe encodings. We list the top-3 frequent discourse markers for specific latent codes that account for \textbf{explicit} discourse relations in the text.}
    \label{fig:case}
\end{figure*}

\begin{figure}[t!]
    \centering
    \includegraphics[width=1.0\columnwidth]{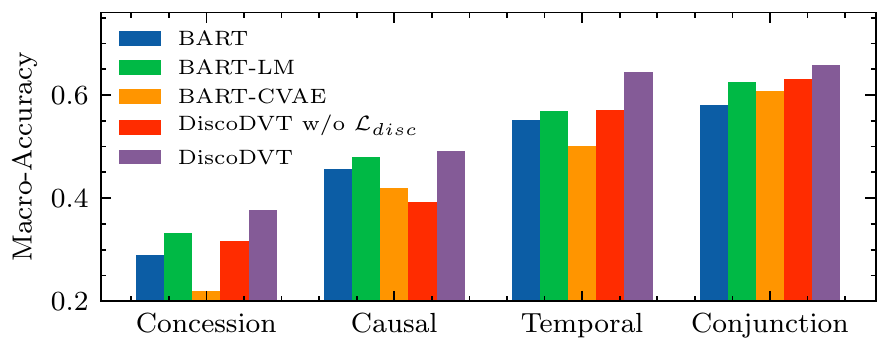}
    \caption{Macro-accuracy of classifying text pairs extracted from the passages generated by different models under the four categories of discourse relations.}
    \label{fig:discourse_coherence}
\end{figure}

\subsection{Discourse-Level Evaluation}\label{sec:disc_eval}

We conduct a comprehensive evaluation of the generated texts on the discourse level. To evaluate the discourse coherence, we extract text pairs connected by discourse markers from the generated texts and train a classifier to predict the relations. 
We then analyze the distribution of discourse relations and show that \textsc{DiscoDVT} uses more diverse discourse patterns as appeared in human texts.


We first fine-tune a BERT model\footnote{We fine-tune on the complete set for one epoch achieving 77.4\% test accuracy that matches the number reported by \citet{DisSent}.} on a dataset
for discourse marker classification~\citep{DisSent} and then train on text pairs extracted from Wikiplots to bridge the domain gap. 


We manually group the discourse markers in $\mathcal{D}$ into four categories based on their most frequent senses~\citep{pdtb2}. Because of the severe class imbalance within each group, we use macro-accuracy.
The results are presented in Figure \ref{fig:discourse_coherence}. We show that \textsc{DiscoDVT} achieves higher accuracy on each category than baselines, especially on temporal relations. We also notice a minor improvement over BART-LM on causal relations since these relations are systematically harder for the model to predict~\citep{DisSent}. Finally, we show the effectiveness of discourse relation modeling by the noticeable accuracy drop when ablating $\mathcal{L}_{disc}$.
We present more details and analysis in the \S{\ref{sec:discourse_coherence}}.

To analyze the distribution of discourse relations, we first show the KL divergence between the distribution generated by a model and that by the human in Table \ref{tab:disc_kl}. \textsc{DiscoDVT} achieves the lowest KL, indicating that the latent codes effectively learn the discourse structure of human texts. We further demonstrate that \textsc{DiscoDVT} can generate more diverse discourse relations in the \S{\ref{sec:discourse_diversity}}.


\begin{table}[t!]
    \centering
    \scriptsize
    \begin{tabular}{c | cccc}
    \toprule[1pt]
    Models & BART & BART-LM & BART-CVAE & \textsc{DiscoDVT} \\
    \midrule[0.5pt]
    KLD $\downarrow$ & 0.0308 & 0.0364 & 0.1700 & \textbf{0.0032} \\
    \bottomrule[1pt]
    \end{tabular}
    \caption{KL divergence (KLD) between the discourse relation distribution by a model and that by the human.}
    \vspace{-10pt}
    \label{tab:disc_kl}
\end{table}

\subsection{Codes Study}

To analyze the correspondence between discrete latent codes and texts, we present a generated example of \textsc{DiscoDVT} with latent codes in Figure \ref{fig:case}. We intuitively assign each discrete latent code to the continuous bpe encodings of length 8, which match the scaling ratio of the CNN. We highlight the discourse markers in each text segment and analyze the corresponding latent code on the right. We list the top-3 frequent discourse markers for each latent code with percentage\footnote{For each latent code, the discourse markers are extracted from 4-grams
that repeat at least two times on the test set.}. We can see that the latent codes learn meaningful correspondence to the discourse relations that guide the model to generate coherent and logical texts. Besides, we also discover that some latent codes learn patterns indicating the beginning or ending of the story, e.g., code ID 216's most frequent 4-gram pattern is \textit{The story is about/set/based}. More generation examples of different models are provided in the \S{\ref{sec:gen_example}}.

\section{Ethics Statement}

We observe that the proposed model may sometimes generate inaccurate or fictitious contents due to the systematic biases of model pre-training on the web corpora and the open-domain characteristics of the story generation datasets. We recommend the users to carefully examine the ethical ramifications of the generated contents in the real-world applications and demonstrations.

\section{Conclusion}

We present \textsc{DiscoDVT}, a discourse-aware discrete variational Transformer for long text generation. \textsc{DiscoDVT} learns a discrete variable sequence that summarizes the global structure of the text, which is then applied to guide the step-wise decoding process to maintain a coherent discourse structure. We further introduce a discourse-aware objective to the discrete latent representations to model discourse relations within the text. Extensive experiments demonstrate that the \textsc{DiscoDVT} can generate long texts with better long-range coherence with interpretable latent codes.

\section*{Acknowledgments}

This work was supported by the National Science Foundation for Distinguished Young Scholars (with No. 62125604) and the NSFC projects (Key project with No. 61936010 and regular project with No. 61876096). This work was also supported by the Guoqiang Institute of Tsinghua University, with Grant No. 2019GQG1 and 2020GQG0005.

\bibliography{anthology,custom}
\bibliographystyle{acl_natbib}


\clearpage
\appendix

\section{Appendices}

\subsection{Ablation Study on CNN Layers}\label{sec:cnn_ablation}

\begin{figure}[h!]
    \centering
    \includegraphics[width=\columnwidth]{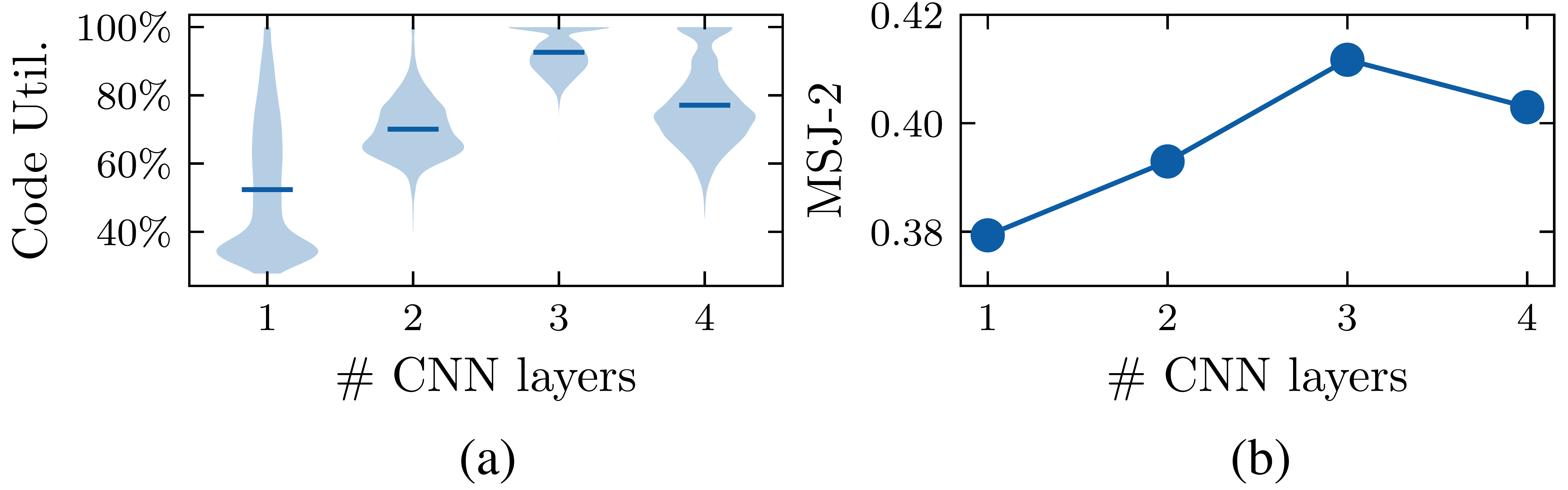}
    \caption{Results of our model with a different number of CNN layers. (a) Distribution of code utilization. (b) Generation performance in MSJ-2.}
    \label{fig:cnn_ablation}
\end{figure}

Since the CNN layers are essential structures in our model for abstracting high-level features of the text, we conduct an ablation study to see the effect of varying the number of CNN layers. Figure \ref{fig:cnn_ablation} (a) plot the distribution of code utilization of the generated examples when using different number of CNN layers. The code utilization is calculated as the type number of used latent codes divided by the length of the latent codes. A high code utilization reflects a sufficient utilization of the whole information capacity of the variational bottleneck where each latent code learns more meaningful information of distinct patterns in the text~\citep{DBLP:conf/icml/KaiserBRVPUS18}.

We observe that the code utilization is maximized when using 3 CNN layers, while either increasing or decreasing the number of CNN layers leads to a decline in the code utilization. We conjecture that when using shallow CNN layers, each latent code only captures local text features and cannot learn high-level information for long text reconstruction. While when using more CNN layers, a larger receptive field for each latent code increases its modeling complexity for longer text chunks. As shown in Figure \ref{fig:cnn_ablation} (b), we present the generation performance in MSJ-2 and observe a similar tendency to the results of the average code utilization.

\subsection{Details on Discourse-Level Evaluation}

\subsubsection{More Analysis on Discourse Coherence}\label{sec:discourse_coherence}

We present fine-grained accuracies for different discourse markers under the four high-level categories in Table \ref{tab:acc_markers}. We observe that our proposed \textsc{DiscoDVT} achieves the highest accuracy on six out of ten classes of discourse markers comparing to the chosen baselines. 
Moreover, \textsc{DiscoDVT}'s performance is more balanced across different discourse markers, while baseline models have low accuracy in specific types of discourse markers, e.g., BART achieves 30\% accuracy on \textit{before}. Finally, we show that even without the discourse-aware objective, the proposed discrete bottleneck learns to abstract some commonly used discourse relations and achieves the highest accuracy on \textit{and} and \textit{also}.

\begin{table*}[t!]
    \centering
    \small
    \begin{tabular}{c|c|cc|cccc|ccc}
    \toprule[1pt]
    \multirow{2}{*}{Model} & \multicolumn{1}{c}{Concession} & \multicolumn{2}{c}{Causal} & \multicolumn{4}{c}{Temporal} & \multicolumn{3}{c}{Conjunction} \\
    \cmidrule(lr){2-2} \cmidrule(lr){3-4} \cmidrule(lr){5-8} \cmidrule(lr){9-11}
     & although & because & so & before & after & as & then & and & also & still \\
     \midrule[0.5pt]
    BART & 0.290 & 0.463 & 0.450 & 0.300 & 0.615 & \textbf{0.564} & 0.730 & 0.839 & 0.511 & 0.389\\
    BART-LM & 0.333 & 0.501 & \textbf{0.460} & 0.500 & 0.620 & 0.486 & 0.667 & 0.844 & 0.529 & 0.500 \\
    BART-CVAE & 0.220 & 0.412 & 0.429 & 0.375 & 0.529 & 0.377 & 0.721 & 0.755 & 0.585 & 0.483 \\
    \midrule[0.5pt]
    \textsc{DiscoDVT} & \textbf{0.377} & \textbf{0.548} & 0.435 & \textbf{0.674} & \textbf{0.623} & 0.547 & \textbf{0.731} & 0.842 & 0.580 & \textbf{0.550} \\
    \ w/o $\mathcal{L}_{disc}$ & 0.317 & 0.361 & 0.425 & 0.458 & 0.612 & 0.549 & 0.667 & \textbf{0.846} & \textbf{0.612} & 0.434 \\
    \bottomrule[1pt]
    \end{tabular}
    \caption{Fine-grained classification accuracies for different discourse markers under four high-level categories.}
    \label{tab:acc_markers}
\end{table*}

\subsubsection{Evaluating Discourse-Level Diversity}\label{sec:discourse_diversity}

To quantitatively understand the discourse diversity of the generated texts, we propose to assess the diversity of discourse relations in the generated texts. 

\begin{figure}[h!]
    \centering
    \includegraphics[width=0.9\columnwidth]{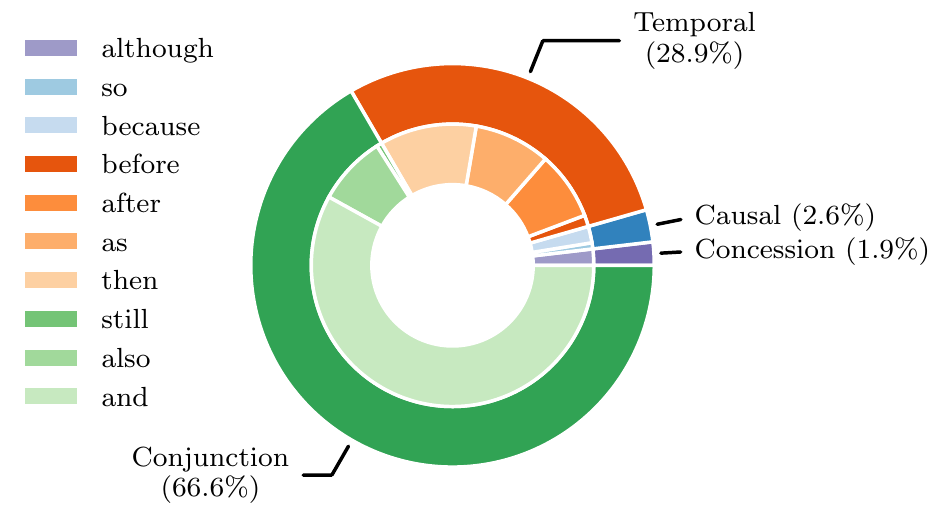}
    \caption{Percentage of different discourse markers and high-level discourse relations in the generated texts of \textsc{DiscoDVT} on Wikiplots.}
    \label{fig:discourse_diversity}
\end{figure}

We first calculate the proportion of different discourse relations and discourse markers used in the texts generated by \textsc{DiscoDVT} on Wikiplots and show the results in Figure \ref{fig:discourse_diversity}. The model shows a diverse preference for different discourse relations that enrich the discourse structure of the generated passages. 


We further compare the utilization percentage of discourse relations across different models and show the results in Table \ref{tab:discourse_diversity}. \textsc{DiscoDVT} exhibits more discourse diversity than the other baselines indicated by a higher entropy score and resembles the golden distribution most closely. Specifically, BART and BART-CVAE mainly generate commonly used discourse markers, such as \textit{and} in conjunction, while express less other complicated relations, such as temporal and causal relations. When ablating the discourse relation modeling objective, we also observe a decline in discourse diversity.

\begin{table}[t!]
    \centering
    \scriptsize
    \begin{tabular}{l | cccc | c}
    \toprule[1pt]
    Models & Conc. & Caus. & Temp. & Conj. & Entr. $\uparrow$ \\
    \midrule[0.5pt]
    Golden & 2.28\% & 2.56\% & 29.06\% & 66.10\% & 1.17 \\
    \midrule[0.5pt]
    BART & 0.84\% & 1.66\% & 23.37\% & 74.13\% & 0.97 \\
    BART-LM & 0.56\% & 1.66\% & 24.30\% & 73.49\% & 0.96 \\
    BART-CVAE & 0.20\% & 1.70\% & 14.07\% & 84.02\% & 0.73 \\
    \midrule[0.5pt]
    \textsc{DiscoDVT} & 1.86\% & 2.59\% & 28.94\% & 66.61\% & \textbf{1.15} \\
    \ w/o $\mathcal{L}_{\text{disc}}$ & 1.74\% & 2.10\% & 27.44\% & 68.73\% & 1.10 \\
    \bottomrule[1pt]
    \end{tabular}
    \caption{Percentage of four categories of discourse relations, i.e., \textbf{Conc}ession, \textbf{Caus}al, \textbf{Temp}oral and \textbf{Conj}unction in the generated samples of different models and the golden references. The \textbf{Entr}opy is calculated over these categories in bits.}
    \label{tab:discourse_diversity}
\end{table}

\subsection{Experimental Details}

\subsubsection{Data Preprocessing}\label{sec:data_proc}

\noindent\textbf{BookCorpus}: We collect a subset of 322k contiguous text segments with at least 512 bpe subwords from the BookCorpus. We reserve the first 512 subwords of each example for training.

\noindent\textbf{Wikiplots}: We use the official split of Wikiplots. We preserve a maximum number of 16 subwords for the input title and 512 subwords for the story, respectively.

\noindent\textbf{WritingPrompts}: We use the official split of WritingPrompts. We strip the accents and the newline markers. We preserve a maximum number of 64 subwords for the input prompt and 512 subwords for the story, respectively.

\subsubsection{Details on Discourse Annotations Preparation}\label{sec:annotation}

We focus on a subset of unambiguous discourse markers $\mathcal{D}$ including \textit{although, so, because, before, after, as, then, and, also} and \textit{still}. An instance of discourse annotation consists of a discourse connective linking a pair of arguments where the first argument (Arg1) is the main clause and the second argument (Arg2) is syntactically bound to the connective.

\citet{pdtb2} found that 61\% of discourse markers and the two arguments appear (with quite flexible order) in the same sentence, and 30\% link one argument to the immediately previous sentence. Due to their high coverage, we focus on automatically extracting these two types of discourse patterns.

\begin{figure*}[h!]
\centering
\resizebox{1.5\columnwidth}{!}{
\begin{tikzpicture}
    \node[] (E) at (-1.0,0) {(a)};
    \node[] (A) at (0,0) {\underline{Although}};
    \node[] (B) at (3.8,0) {$[\textbf{61 billion people have perished,}]_{\text{Arg2}}$};
    \node[] (C) at (9.7,0) {$[\textit{Paul's prescient visions indicate that}$};
    \node[] (D) at (4, -0.8) {$\textit{this is far from the worst possible outcome for humanity.}]_{\text{Arg1}}$};
    \draw[->, line width=0.5pt] (B) to [bend right=30] node[above] {mark} (A);
    \draw[->, line width=0.5pt] (C) to [bend right=30] node[above] {advcl} (B);
\end{tikzpicture}
}

\resizebox{1.55\columnwidth}{!}{
\begin{tikzpicture}
    \node[] (E) at (-4.2,0) {(b)};
    \node[] (A) at (0,0) {$[\textit{Father Matthew decides to send him to Rome,}]_{\text{Arg1}}$};
    \node[] (B) at (4.2,0) { \underline{so}};
    \node[] (C) at (7.1,0) {$[\textbf{he can attend an exorcism class}$};
    \node[] (D) at (-1.8, -0.8) {$\textbf{taught by his friend}]_{\text{Arg2}}$};
    \draw[->, line width=0.5pt] (A) to [bend left=20] node[above] {advcl} (C);
    \draw[->, line width=0.5pt] (C) to [bend left=30] node[below] {mark} (B);
\end{tikzpicture}
}

\resizebox{1.6\columnwidth}{!}{
\begin{tikzpicture}
    \node[] (E) at (-1.0,0) {(c)};
    \node [] (A) at (0,0) {\underline{Because}};
    \node [] (B) at (3.7,0) {$[\textbf{the detectives do not believe her}]_{\text{Arg2}}$};
    \node [] (C) at (10.2,0) {$[\textit{, she decides to contact Gerard herself.}]_{\text{Arg1}}$};
    \draw[->, line width=0.5pt] (B) to [bend right=30] node[above] {mark} (A);
    \draw[->, line width=0.5pt] (C) to [bend right=30] node[above] {advcl} (B);
\end{tikzpicture}
}

\resizebox{1.57\columnwidth}{!}{
\begin{tikzpicture}
    \node[] (E) at (-2.7,0) {(d)};
    \node [] (A) at (0,0) {$[\textit{He damages the stabilizer}]_{\text{Arg1}}$};
    \node [] (B) at (3.0,0) {\underline{before}};
    \node [] (C) at (7.6,0) {$[\textbf{his teammates can tie him up in the shuttle.}]_{\text{Arg2}}$};
    \draw[->, line width=0.5pt] (C) to [bend right=25] node[above] {mark} (B);
    \draw[->, line width=0.5pt] (A) to [bend left=35] node[above] {advcl} (C);
\end{tikzpicture}
}

\resizebox{1.5\columnwidth}{!}{
\begin{tikzpicture}
    \node[] (E) at (-0.7,0) {(e)};
    \node [] (A) at (0,0) {\underline{After}};
    \node [] (B) at (2.7,0) {$[\textbf{Cho calms him down,}]_{\text{Arg2}}$};
    \node [] (C) at (8.8,0) {$[\textit{he follows the captain's order to fix the drive.}]_{\text{Arg1}}   $};
    \draw[->, line width=0.5pt] (B) to [bend right=40] node[above] {mark} (A);
    \draw[->, line width=0.5pt] (C) to [bend right=30] node[above] {advcl} (B);
\end{tikzpicture}
}

\resizebox{1.3\columnwidth}{!}{
\begin{tikzpicture}
    \node[] (E) at (-0.5,0) {(f)};
    \node [] (A) at (0,0) {\underline{As}};
    \node [] (B) at (2.1,0) {$[\textbf{his powers drain,}]_{\text{Arg2}}$};
    \node [] (C) at (7.6,0) {$[\textit{Luthor wishes the experience to continue.}]_{\text{Arg1}}$};
    \draw[->, line width=0.5pt] (B) to [bend right=40] node[above] {mark} (A);
    \draw[->, line width=0.5pt] (C) to [bend right=30] node[above] {advcl} (B);
\end{tikzpicture}
}

\resizebox{1.2\columnwidth}{!}{
\begin{tikzpicture}
    \node[] (E) at (-2.7,0) {(g)};
    \node [] (A) at (0,0) {$[\textit{Mason destroys the chips,}]_{\text{Arg1}}$};
    \node [] (B) at (2.8,0) {\underline{then}};
    \node [] (C) at (5.5,0) {$[\textbf{surrenders to Hummel.}]_{\text{Arg2}}$};
    \draw[->, line width=0.5pt] (A) to [bend left=20] node[above] {parataxis} (C);
    \draw[->, line width=0.5pt] (C) to [bend left=30] node[below] {advmod} (B);
\end{tikzpicture}
}

\resizebox{1.5\columnwidth}{!}{
\begin{tikzpicture}
    \node[] (E) at (-4.8,0) {(h)};
    \node [] (A) at (0,0) {$[\textit{Nick blames Jerry for forcing him into the profession}]_{\text{Arg1}}$};
    \node [] (B) at (4.8,0) {\underline{and}};
    \node [] (C) at (7.3,0) {$[\textbf{asks him to get away.}]_{\text{Arg2}}$};
    \draw[->, line width=0.5pt] (A) to [bend left=20] node[above] {conj} (C);
    \draw[->, line width=0.5pt] (C) to [bend left=40] node[below] {cc} (B);
\end{tikzpicture}
}

\resizebox{1.42\columnwidth}{!}{
\begin{tikzpicture}
    \node[] (E) at (-5.1, 0) {(i)};
    \node[] (A) at (0,0) {$[\textit{Kenny, revealed to be alive and an undercover FBI agent.}]_{\text{Arg1}}$};
    \node[] (B) at (6.5,0) {$[\textbf{He \underline{also} implies that}$};
    \node[] (C) at (-1.2,-0.8) {$\textbf{Lampone is another undercover agent.}]_{\text{Arg2}}$};
    \draw[<-, line width=0.5pt] (B) to [bend right=20] node[above] {next sentence*} (A);
\end{tikzpicture}
}

\resizebox{1.38\columnwidth}{!}{
\begin{tikzpicture}
    \node [] (E) at (-4.6,0) {(j)};
    \node [] (A) at (0,0) {$[\textit{She strikes out across the dense sawgrass marshes}]_{\text{Arg1}}$};
    \node [] (B) at (4.6,0) {\underline{still}};
    \node [] (C) at (6.8,0) {$[\textbf{miles from home.}]_{\text{Arg2}}$};
    \draw[->, line width=0.5pt] (C) to [bend left=40] node[below] {advmod} (B);
    \draw[->, line width=0.5pt] (A) to [bend left=20] node[above] {dep} (C);
\end{tikzpicture}
}

    \caption{Dependency patterns of the 10 discourse markers in $\mathcal{D}$ with annotated examples from Wikiplots. Labels above the arrows are grammatical relations defined in \citet{manning-EtAl:2014:P14-5}.*: The \textit{next sentence} pattern identifies the adjacent two sentences as Arg1 and Arg2. }
    \label{fig:patterns}
\end{figure*}
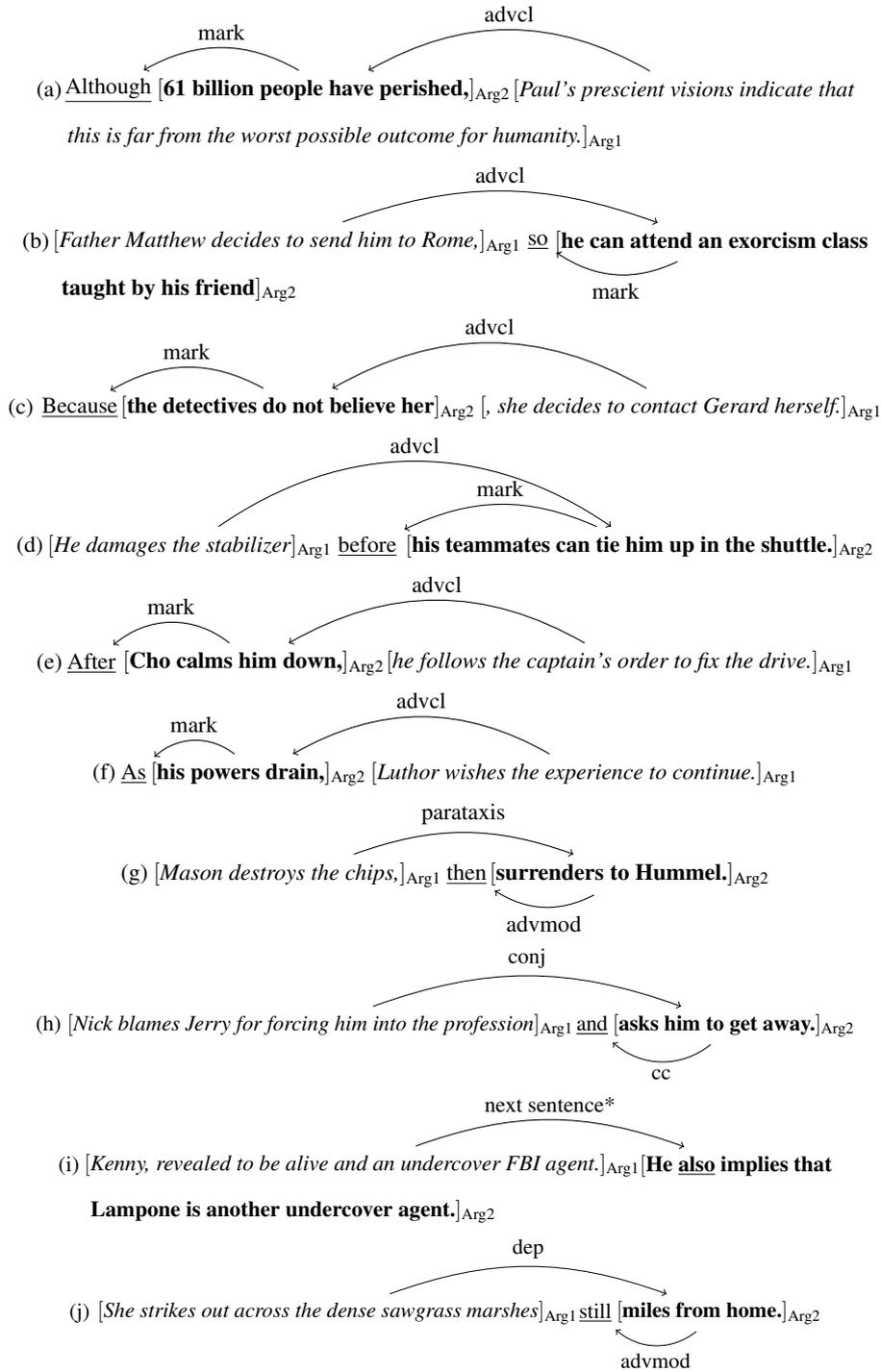

We resort to universal dependency grammar that provides sufficient information to extract the discourse markers and the associated two arguments. For each discourse marker of interest, we follow \citet{DisSent} and use appropriate dependency patterns to extract intra-sentence discourse relations as shown in Figure \ref{fig:patterns}. In each example, Arg1 is in \textit{italics}, Arg2 is in \textbf{boldface}, and the discourse marker is \underline{underlined}.

We first parse each sentence into a dependency tree with the Stanford CoreNLP toolkit~\citep{manning-EtAl:2014:P14-5}. Then we identify discourse markers and the spans of Arg1 and Arg2 based on the dependency patterns and ensure that the two arguments together with the marker cover the whole sentence. If there are multiple discourse markers in one sentence, we preserve the one that divides the sentence more evenly.
If the parsing results reveal that there is only one argument in the sentence that connects with the discourse marker, we heuristically label the previous adjacent sentence as Arg1 (see the \textit{next sentence} pattern in Figure \ref{fig:patterns}).

For each story in the dataset, we first split it into individual sentences and then apply the above steps for extraction until all the adjacent EDUs (either a complete sentence or a parsed sub-sentence) are labeled with proper relations. The label candidates are the combination of discourse markers from $\mathcal{D}$ and two possible directions that indicate how these two text spans are linked together. For example, in Figure \ref{fig:patterns} (a), the label of the text pair will be \texttt{although\_arg2\_arg1} according to the order of the two arguments. If no discourse relation is identified from the pair of text spans, they are labeled with \textit{unknown}.

\subsubsection{Details on Training Settings}\label{sec:train_detail}

To improve the reproducibility of our model, we provide the detailed training settings in this section. 

We implement our codes based on the repository of Huggingface's Transformers~\citep{wolf-etal-2020-transformers}. For the posterior network, we initialize the Transformer encoder with the pre-trained parameters of the encoder of $\text{BART}_{\text{base}}$ (82M parameters). The generator model is initialized with the pre-trained checkpoint of $\text{BART}_{\text{base}}$ (140M parameters). Other randomly initialized parameters, including the CNN layers, the transposed CNN layers, the latent code embeddings, etc., sum up to 2.5M parameters. The prior network is also a Transformer encoder-decoder that uses the same architecture as $\text{BART}_{\text{base}}$ (140M parameters).

\noindent\textbf{Warm-Start Training}

We collect 322K contiguous texts from BookCorpus~\citep{bookcorpus} and keep the first 512 bpe subwords of each example for training. Since the warm-start training aims at initializing the latent embeddings for reconstructing the target text, we do not feed any input to the encoder. We use a fixed Gumbel temperature of 0.9 and a fixed learning rate of 1e-4. We use a batch size of 4 and a gradient accumulation step of 4 and train on the collected data for one epoch which takes about 7 hours on 1 GeForce RTX 2080 (11G).

\noindent\textbf{Fine-tuning}

For fine-tuning the generator and the posterior network for text reconstruction, we anneal the Gumbel temperature from $\tau_{max}=0.9$ to $\tau_{min}=0.1$ using exponential decay schedule where the Gumbel temperature $\tau$ at step $T$ is: $\max[\tau_{min}, \tau_{max}\times \exp(-10^{-4}\times T)]$. We linearly decrease the learning rate from 1e-4 to 0 throughout the fine-tuning. We use a batch size of 4 and a gradient accumulation step of 4. We fine-tune for five epochs on Wikiplots and one epoch on WritingPrompts, which takes about 12 hours and 6 hours on 1 GeForce RTX 2080 (11G), respectively. 

For fine-tuning the prior network, we initialize the encoder with the pre-trained parameters of the encoder of $\text{BART}_{\text{base}}$. We linearly decrease the learning rate from 1e-4 to 0 during training. The maximum target sequence length is set to $\text{MaxLength}=64$. We use a batch size of 128 and a gradient accumulation step of 8. We fine-tune the model for 100 epochs which takes about 13 hours on 1 GeForce RTX 2080 (11G). During inference, we randomly sample a sequence of latent codes from the prior network autoregressively and set the minimum sequence length to 38 and 44 for Wikiplots and WritingPrompts, respectively.

We present the hyper-parameter search space in Table \ref{tab:finetunesearch}. The number of hyper-parameter search trials was 10. We adopt a manual search to select the hyper-parameters, and the selection criterion was BLEU-1 on the validation set when fine-tuning on the downstream datasets.

\begin{table} [!htp]
\centering
\small
\setlength{\tabcolsep}{0.5mm}{
\begin{tabular}{cc}
\toprule[1pt]
Hyper-parameter & Search Space \\
\midrule[0.5pt]
Learning Rate & \textit{choice}[8e-5, 1e-4, 2e-4] \\
Training Epoch & \textit{choice}[1, 2, 3, 4, 5] \\
Batch Size & \textit{choice}[4, 8, 16] \\
Input Length & \textit{choice}[16, 32, 64] \\
Output Length & 512 \\
Latent Vocabulary Size & \textit{choice}[32, 64, 128, 256, 512] \\
Top-p & \textit{choice}[0.8, 0.9, 1.0] \\
Maximum Gradient Norm & 1.0 \\
Optimizer & AdamW \\
Epsilon (for AdamW) & 1e-8 \\
\bottomrule[1pt]
\end{tabular}}
\caption{Hyper-parameter search space of \textsc{DiscoDVT} during fine-tuning. \textit{choice} indicates that the listed numbers will be chosen with the same probability.}
\label{tab:finetunesearch}
\end{table}

\section{Generation Examples}\label{sec:gen_example}

We present the generation examples from different models on Wikiplots in Table \ref{tab:gen_example}. We observe that baselines including BART-LM and BART-CVAE suffer from repetition problem as the contexts grow longer while \textsc{DiscoDVT} maintains a better discourse structure of the text by the global discrete latent codes and stays on topic throughout the whole story (indicated by the frequent usage of the keyword ``\textit{Probe 7}''). 
Finally, we observe that existing models have potential spelling mistakes when generating complicated entities in the open-domain scenario. This phenomenon is severe in BART as in the example, which even hinders readability. We suggest that a future direction to mitigate this issue is to construct entity plots by substituting entity mentions with placeholders and apply the copy mechanism.

\begin{table*}[t!]
    \centering
    \small
    \begin{tabularx}{2\columnwidth}{l|X}
    \toprule[1pt]
    Title & \textit{Probe 7, Over and Out} \\
    \midrule[0.5pt]
    {BART} & In a field hospital, dr David \underline{Mihalya} discovers a corpse in a nearby car. 
His search leads him to a facility where \textcolor{purple}{his own corpse is being guarded by a secret military unit.} 
\underline{Mihalai} visits the unit and discovers that the building is a military barracks built in 1940, and that a number of the soldiers are killed in an explosion. 
\underline{Muhalya} is taken to the military base to be interrogated, but is brought into conflict with his commanding officer, Captain James \underline{Mihalia}, who also appears to be involved. 
\underline{Mohalya} asks \underline{Mihali} to give him a call as a replacement for \underline{Mihail}. 
\underline{Mishali} says that he will be sent to investigate, and he goes to see \underline{Mihanya}. 
\underline{Muthali} says they were sent to an undisclosed location when he was a child, and they are now in the same military camp. 
\underline{Mighali} and \underline{Mihaela} begin to see each other, and the two are seen to have a casual sex. 
\underline{Makhalya} says she was "happy about her" as \underline{Mihiya} said he never saw her. 
\underline{Mumbali} is shocked at \underline{Mihmya}'s response and tries to get her to get to know \underline{Mihyasa}, but she rejects him. \\
    \midrule[0.5pt]
    {BART-LM} & In a field where men are in labor and \textcolor{purple}{women are selling women}, a research firm is in session with an eccentric scientist who is trying to determine a cure for a blood disorder. 
The scientist goes to a hospital to visit the woman and discovers that her liver is growing to a dangerously high size, and she has a blood clot. 
The technician explains that she has to be careful not to leave the woman to go through an experimental procedure to remove the clot, but she is unable to live with the strain and her liver can't survive without a transplant. 
After the procedure, the team finds that the woman's liver has enlarged to a lethal size. 
However, in the process, the patient gets out of the hospital and is sent to a large hospital in the countryside where he meets with a young woman named Elizabeth. \textcolor{red}{Elizabeth does not recognise Elizabeth, but Elizabeth quickly recognises Elizabeth's condition. Elizabeth begins to suspect Elizabeth and Elizabeth's secret relationship and believes Elizabeth to be linked to Elizabeth.}\\
    \midrule[0.5pt]
    {BART-CVAE} & The series tells the story of Peter Smith, a boy from a small town, who fell in love, and decided to find the way to end the romance. 
Peter and his brother and friend, Mike, a successful lawyer, return to their town, and have the prospect of dating and marrying. 
After Peter discovers that Mike is with Paul, he decides to take his sister, Laura, with him and give them and his friends the word "love" and so on, until his sister says "Yes". 
It is up to Peter to resolve this dispute, but to reach an end with the men as good-luck, and help people in other ways. 
After a short visit to the church, Paul discovers that Pete and Laura are a couple, and has made a difference of how people think. 
He becomes concerned and asks Laura about Peter, so as to make her believe that she is, in fact, married to Peter. 
In order to resolve the situation, Peter uses a combination of the words "Love" and "Dollar" (for "love") in the name of a woman, but, for Peter, he can only say the words at the end and, when Peter, Laura and Peter end up in a bed in Peter's home, as far as Peter sees, is lying on the bed. 
Peter thinks about what to do at the hospital, and Laura tells him to go look at Pete. 
\textcolor{red}{Peter asks Peter about Pete and goes}, and Pete mentions that David's wife is being married with two children, but Peter says that the children can not be married, or that he isn't with her for such a short period of time. 
\textcolor{red}{Peter goes over to Peter's house, and he asks Peter for a second time.} \\
    \midrule[0.5pt]
    \textsc{DiscoDVT} & The story begins in the near future in the year 2009. 
An alien race called the Invaders appear and break the Galactic Federation and destroy the Cardassians from their planet. 
A group of mercenaries called the "Blue Angels of New Generation" are tasked with trying to destroy the Defiant. 
They use a device called a \textbf{Probe 7} in order to kill the Invaders. 
The \textbf{probe} must send a distress signal to the Federation starship, commanded by the "Sister's" pilot, Professor \underline{Moriarty} (voiced by Arthur Fairchild from the film The Secret Intelligence Service). 
Professor \underline{Moriory}, along with the Enterprise, arrive and successfully intercept the Invaders while the ship remains on orbit and attacks the Federation Fleet on the planet. 
The Blue Angels then infiltrate the fleet to set a trap for the Invaders, hoping that they will destroy the fleet. 
As the Blue Angels use this technology, the Red Angels of The Invaders retaliate by destroying the Enterprise before they reach the Federation fleet. 
The Invaders then proceed to destroy all of Earth's radio stations and fire the \textbf{Probe 7} into the "Dominic Channel". 
Professor \underline{Moriorthy} then uses \textbf{Probe 7} to gain access to his ship's central control, which contains an orbiting outpost called the Black Mesa. \\
    \bottomrule[1pt]
    \end{tabularx}
    \caption{We present the first ten sentences of the stories generated by different models. We highlight the obvious \textcolor{red}{repetitions}, \textcolor{purple}{unreasonable descriptions}, and potential \underline{spelling mistakes} in the generated stories. The generated keywords that match the title are presented in \textbf{boldface}.}
    \label{tab:gen_example}
\end{table*}



\end{document}